\documentclass{article}

\usepackage{microtype}
\usepackage{graphicx}
\usepackage{subcaption}
\usepackage{booktabs} % for professional tables

\usepackage{hyperref}

\usepackage[preprint]{icml2026}

\usepackage{amsmath}
\usepackage{amssymb}
\usepackage{mathtools}
\usepackage{amsthm}

\usepackage[capitalize,noabbrev]{cleveref}

\theoremstyle{plain}

\theoremstyle{definition}

\theoremstyle{remark}

\usepackage[textsize=tiny]{todonotes}

\icmltitlerunning{Aggregating Direct and Indirect Neighbors through Graph Linear Transformations}

\begin{document}

\twocolumn[
\icmltitle{Aggregating Direct and Indirect Neighbors through \\
            Graph Linear Transformations}



  \icmlsetsymbol{equal}{*}

  \begin{icmlauthorlist}
    \icmlauthor{Marshall Rosenhoover}{yyy}
    \icmlauthor{Huaming Zhang}{yyy}
  \end{icmlauthorlist}

    \icmlaffiliation{yyy}{Department of Computer Science, University of Alabama in Huntsville, Huntsville, Alabama, United States of America}

\icmlcorrespondingauthor{Marshall Rosenhoover}{marshall.rosenhoover@uah.edu}

  \icmlkeywords{Graph Linear Transformation, Gaussian Belief Propagation, Gaussian Graphical Models}

  \vskip 0.3in
]



\printAffiliationsAndNotice{}  % no special notice (required even if empty)

\begin{abstract}
    Graph neural networks (GNN) typically rely on localized message passing, requiring increasing depth to capture long range dependencies. In this work, we introduce Graph Linear Transformations, a linear transformation that realizes direct and indirect feature mixing on graphs through a single, well-defined linear operator derived from the graph structure. By interpreting graphs as walk-summable Gaussian graphical models, we compute these transformations via Gaussian Belief Propagation, enabling each node to aggregate information from both direct and indirect neighbors without explicit enumeration of multi-hop paths. We show that different constructions of the underlying precision matrix induce distinct and interpretable propagation biases, ranging from selective edge-level interactions to uniform structural smoothing, and that Graph Linear Transformations can achieve competitive or superior performance compared to both local message-passing GNNs and dynamic neighborhood aggregation models across homophilic and heterophilic benchmark datasets.
\end{abstract}

\section{Introduction}
    The standard idea behind feed-forward neural networks is to apply a linear transformation to the input features followed by a nonlinear activation function. Applying this concept to graphs, it would be natural to define a graph linear transformation followed by a nonlinear activation function, where the graph linear transformation corresponds to a matrix operator derived from the graph structure acting on node features. Such a transformation would aggregate information from all nodes, weighted by the structure of the graph. However, explicitly constructing such a transformation --- by enumerating the contributions along all multi-hop paths in the graph --- is not feasible, as this requires reasoning over an exponential number of path-dependent interactions, a problem known to be NP-hard ~\cite{sum_of_walks_intractable, sum_of_walks_intractable_2}.

    This intractability is consistent with the practical use of localized, layer-wise approximations that propagate information through the graph in a compositional manner. In particular, Graph Convolutional Networks \cite{GCN} and related message-passing architectures can be viewed as approximations to graph linear transformations by repeatedly aggregating information from local neighborhoods. While effective in many settings, these methods rely on increasing network depth to approximate first order and higher order feature mixing and therefore exhibit an inherent inductive bias toward locality.

    More recent research has gathered toward dynamic neighborhood aggregation in the form of attention-based models and continuous dynamic models to relax the fixed-depth limitations of local layer-wise approximations. However, these methods realize wider context through dynamic composition --- either by stacking local interactions over time or by relaxing neighborhood structure to enable dense interactions --- rather than through a single graph-induced linear transformation that jointly captures direct and indirect dependencies.
    
    In this work, we ask whether graph linear transformations can be realized directly, without relying on deep stacks of local operators. We show that such transformations can be computed by imposing simple structural constraints, and that they can be either fixed or learned from data. Our results suggest that local and nonlocal graph mixing can be achieved directly, rather than through depth-based approximations.

    \begin{figure*}[t] 
        \centering
        \includegraphics[width=0.8\textwidth]{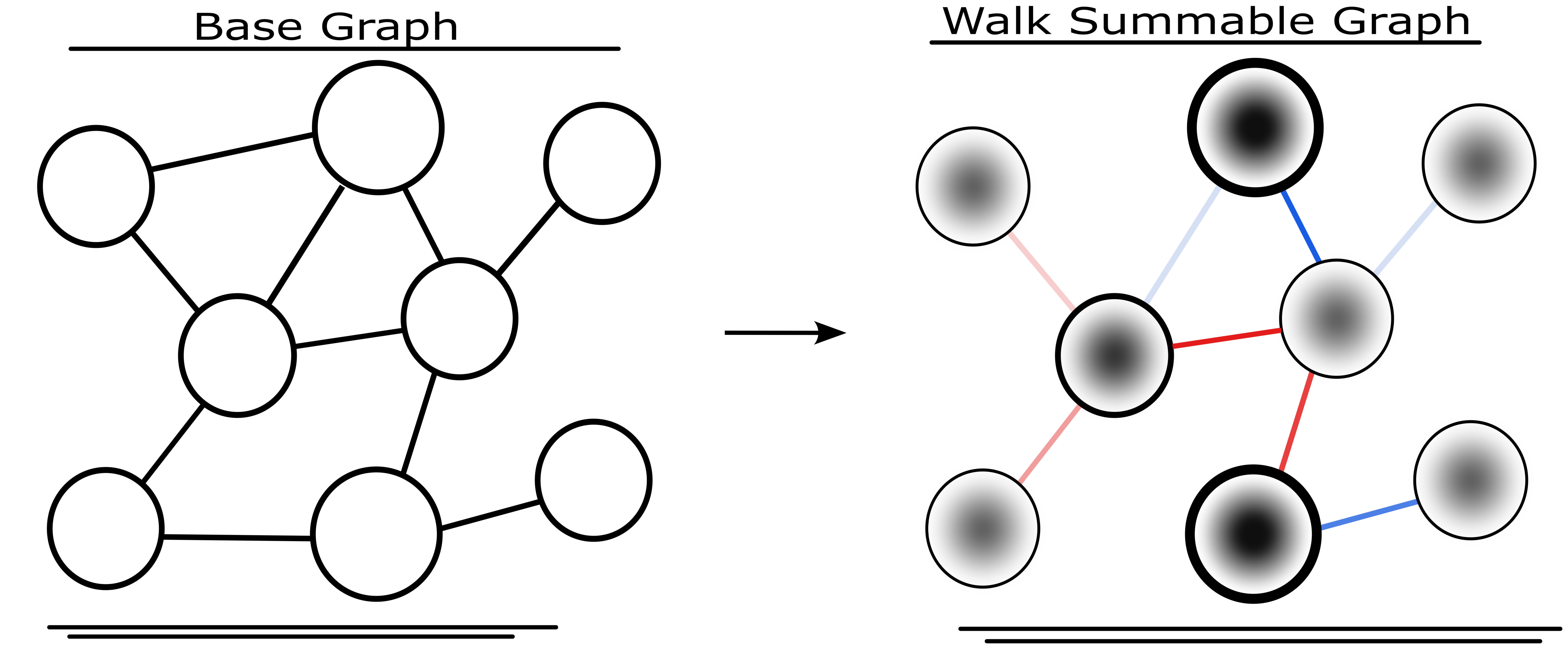}  
        \caption{Illustration of a graph being interpreted as a walk summable Gaussian graphical model, where node shading encodes confidence and edge color denotes interaction polarity and strength.}
        \label{fig:reinterpretation}
    \end{figure*}

%%%%%%%%%%%%%%%%%%%%%%%%%%%%%%%%%%%%%%%%%%

\section{Background}
    Gaussian graphical models represent a set of continuous random variables \( x = [x_1, \ldots, x_N]^\top \) that follow a joint multivariate Gaussian distribution with covariance matrix \( \Sigma \). Each variable is associated with a node in a graph, where edges capture conditional dependencies between variables. The precision matrix \( J = \Sigma^{-1} \) encodes these conditional dependencies with nonzero entries \( J_{ij} \) indicating edges and edge strength between nodes with \( J_{ii} \) encoding the strength of a node's self confidence ~\cite{malioutov2006walk, GaBP_Thesis, GaBP_intro}. Together with this structural information, the model includes an information vector \( h \) representing the nodes' local evidence. The pair \( (J, h) \) fully defines the probabilistic state of the system, with the joint distribution
    \[
    p(x) \propto \exp\!\left(-\tfrac{1}{2}x^\top J x + h^\top x\right),
    \]
    where \( J \in \mathbb{R}^{N \times N} \), \( h \in \mathbb{R}^N \), and \( x \in \mathbb{R}^N \). The equilibrium of this system is thus given by 
    \[
    \mu = J^{-1}h.
    \]
    However, directly inverting \( J \) becomes computationally expensive as graph size scales. Gaussian Belief Propagation (GaBP) addresses this by solving the linear system through local message passing on the graph induced by the sparsity pattern of $J$. In the univariate setting, GaBP associates a scalar message with each directed edge $i \to j$. Each message is parameterized by a precision $\pi_{i \to j}$, current confidence in its prediction, and an information term $\eta_{i \to j}$, the current belief of the node weighted by its belief. At each iteration, node $i$ computes a message to neighbor $j$ using all incoming messages from neighbors other than $j$. Let $\mathcal{N}(i) \setminus j$ denote this set, and define
    \[
        \alpha_{i \setminus j}
        = J_{ii} + \sum_{k \in \mathcal{N}(i) \setminus j} \pi_{k \to i},
        \qquad
        \beta_{i \setminus j}
        = h_i + \sum_{k \in \mathcal{N}(i) \setminus j} \eta_{k \to i}.
    \]
    The message updates are then given by
    \[
        \pi_{i \to j}
        = -\frac{J_{ij}^2}{\alpha_{i \setminus j}},
        \qquad
        \eta_{i \to j}
        = -\frac{J_{ij}}{\alpha_{i \setminus j}} \, \beta_{i \setminus j}.
    \]

    After convergence, the marginal parameters at each node are obtained by aggregating all incoming messages,
    \[
        \pi_i
        = J_{ii} + \sum_{k \in \mathcal{N}(i)} \pi_{k \to i},
        \qquad
        \eta_i
        = h_i + \sum_{k \in \mathcal{N}(i)} \eta_{k \to i},
    \]
    yielding the solution $\mu_i = \eta_i / \pi_i$.
   
    \paragraph{Convergence Guarantees.}
        GaBP is guaranteed to converge on cyclic graphs only when the precision matrix is walk-summable. Under this condition, the total influence of all walks remains finite. Walk-summability, thus, requires that the normalized precision matrix $\tilde{J} = D^{-1/2} J D^{-1/2}$, where $D$ is the diagonal of the precision matrix, satisfies the spectral condition
        \[
            \rho\bigl(|I - \tilde{J}|\bigr) < 1.
        \]

%%%%%%%%%%%%%%%%%%%%%%%%%%%%%%%%%%%%%%%%%%

\section{Graph Linear Transformations}

    \begin{figure}[h]
        \centering
        \includegraphics[width=0.6\columnwidth]{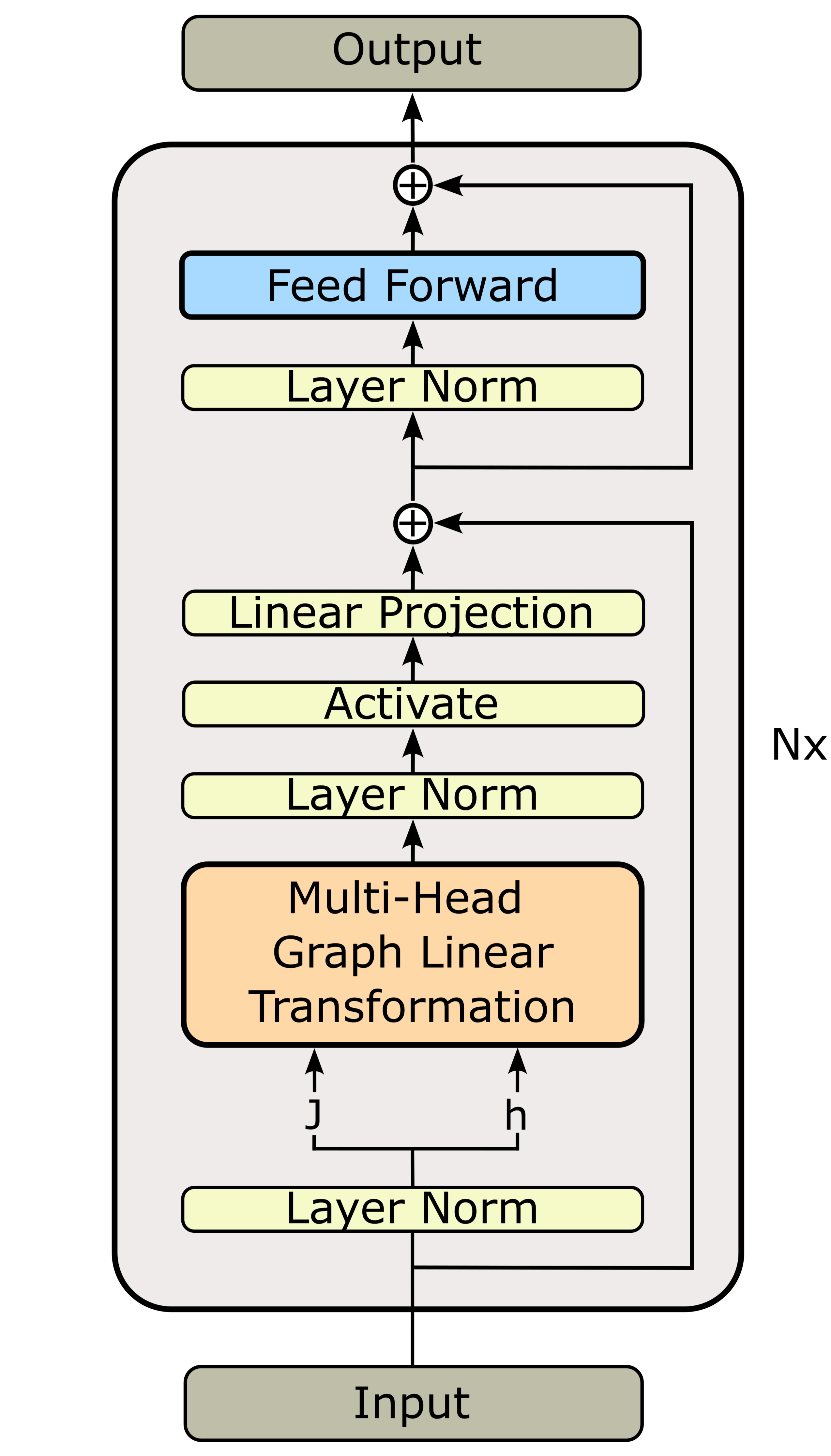}  
        \caption{Architecture of the Graph Linear Transformation Layer.}
        \label{fig:model_architecture}  
    \end{figure}
        
    To realize graph linear transformations directly, we reinterpret graphs as walk-summable Gaussian graphical models, as illustrated in Figure~\ref{fig:reinterpretation}. In this formulation, Gaussian dependencies implicitly accumulate the influence of all paths in the graph, eliminating the need to enumerate them. By parameterizing the node evidence $h$ and, optionally, the topology $J$, we introduce the graph linear transformation layer. In this, the model is able to learn how strongly each node should participate in their neighborhood and structure of the neighborhoods. 

    \subsection{Graph Linear Transformation Layer}
    \label{sec:gabp_architecture}   graphical models
        The overall structure of the Graph Linear Transformation Layer, illustrated in Figure \ref{fig:model_architecture}, consists of two subblocks: a Graph Linear Transformation block followed by a node-wise feed-forward network. Each subblock applies layer normalization to the node features before processing, following the pre-normalization design shown to improve training stability in Transformer architectures \cite{prenorm}. In addition, the Graph Linear Transformation block applies a second layer normalization after the graph linear transformation to mitigate large, graph-dependent scale variations arising from differences in node degrees and edge weights. Since the operation itself is linear, a nonlinear activation function is applied to the updated node embeddings, which are then projected back to the original embedding dimension through a linear layer. Residual connections are incorporated around each subblock, prior to the normalization step. 

    \subsection{Creation of Graph Linear Transformations}
        In GNNs, nodes are typically represented by multidimensional embeddings. Extending GaBP to be multivariate, where each node is multidimensional, causes the messages to be matrices that couple all embedding dimensions. To maintain scalability, we therefore make two simplifying assumptions. First, the neural network maps input features into an embedding space where dimensions can be treated as independent. Second, dimensions share a global precision matrix $J$. Intuitively, this means that each embedding dimension in $h$ can be treated as a univariate case of GaBP and is propagated through the same topology $J$.

        \subsubsection{Precision Matrix Designs}
        \label{sec:matrix_construction}
            Different constructions of \( J \) induce different inductive biases of information propagation, helping to shape whether propagation behaves as local, global, smooth, or selective. We consider three walk-summable designs of \( J \): Pairwise Normal, Diagonally Dominant, and Laplacian. 
            
            \paragraph{Pairwise Normal} 
                Pairwise-normalizable Gaussian graphical models span the same class of precision matrices as walk-summable models~\cite{malioutov2006walk}. Each edge \(e = (i,j)\) contributes a local precision block
                \[
                J^{(e)} =
                  \begin{pmatrix}
                  a_{ij} & b_{ij}\\[2pt]
                  b_{ij} & c_{ij}
                  \end{pmatrix},
                \qquad
                    a_{ij} c_{ij} > b_{ij}^{2},
                \]
                where the terms \(a_{ij}\) and \(c_{ij}\) represent the self-precisions of nodes \(i\) and \(j\), while \(b_{ij}\) encodes their mutual coupling. For the full precision matrix \(J\), each off-diagonal entry corresponds to the edge coupling \(J_{ij} = b_{ij}\), and each diagonal entry aggregates a node’s self-precision contributions from all adjacent edges:
                \[
                J_{ii} = \sum_{j \in \mathcal{N}(i)} a_{ij}.
                \]
                This construction induces a mutual-consistency bias, since the effective strength of an edge depends jointly on both endpoints: edges where either node has low self-confidence in the edge contribute weak couplings, while edges supported by two nodes with a high self-confidence in the edge contribute strong ones. As a result, information propagation is naturally selective, emphasizing well-supported connections and attenuating uncertain or inconsistent ones.
    
            \paragraph{Diagonally Dominant}
                Diagonally dominant matrices are a subset of walk-summable matrices~\cite{GaBP_Thesis}. Such matrices are characterized by having each diagonal entry larger in magnitude than the sum of the magnitudes of the off-diagonal entries in the same row, that is,
                \[
                J_{ii} \;>\; \sum_{j \neq i} |J_{ij}|.
                \]
                In this design, each node is more confident in itself than in its connections to other nodes. This induces a confidence-centered bias, where information tends to flow outward from nodes with large diagonal precision and is absorbed by nodes with low self-precision. As a result, propagation becomes locally anchored, with high-confidence nodes acting as influential hubs and low-confidence nodes behaving primarily as receivers rather than broadcasters.

            \begin{figure*}[t] 
                \centering
                \includegraphics[width=\textwidth]{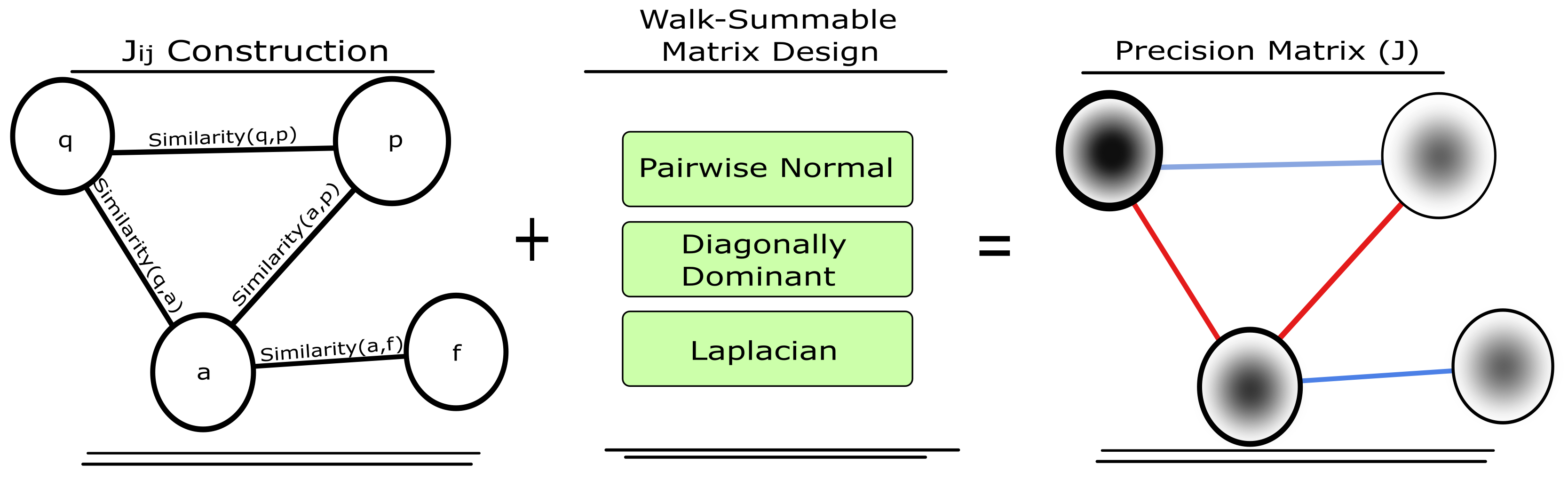}  
                \caption{Depiction of how to construct a precision matrix \(J\).}
                \label{fig:constructionJ}  
            \end{figure*}

            \paragraph{Laplacian}
                Laplacian-based matrices are not inherently walk-summable matrices as their eigenvalues inherently lie in the range $[-1, 1]$. They are derived from the normalized graph Laplacian as
                \[
                L = I - D^{-\frac{1}{2}} A D^{-\frac{1}{2}},
                \]
                where $A$ is the weighted adjacency matrix and $D$ is the corresponding degree matrix. By rescaling $D^{-\frac{1}{2}} A D^{-\frac{1}{2}}$  to lie in the range $(-1, 1)$, the resulting matrix becomes walk-summable. In this, information is smoothed according to the connectivity of the graph structure, inducing a manifold-smoothing bias where nodes embedded in similar structural environments co-vary strongly. As a result, propagation tends to operate at a community or cluster level, promoting broad diffusion over the graph while preserving contrasts between different structural regions.
        
        \subsubsection{Parameterization of Graph Linear Transformations}

            For the learnable components of the graph linear transformation, the observation vector \(h\) and precision matrix \(J\) are parameterized from base node embeddings. The observation vector \(h\) is obtained by projecting the node embeddings
            \(X \in \mathbb{R}^{N \times d_{\text{model}}}\) into a latent observation space,
            \[
                h = \text{LeakyReLU}(X W_{\text{obs}}), \qquad
                W_{\text{obs}} \in \mathbb{R}^{d_{\text{model}} \times d_{\text{latent}}}.
            \]
            
            To construct the precision matrix \(J\), edge-level interactions are derived either from a fixed graph structure or from learned pairwise node similarities. The overall construction process is summarized in Figure~\ref{fig:constructionJ}. In the learned setting, node embeddings are first mapped into a similarity space, either by directly using the observation vector \(h\) to compute similarity-based edge strengths or by learning a separate similarity projection,
            \[
                s = \text{LeakyReLU}(X W_{\text{sim}}), \qquad
                W_{\text{sim}} \in \mathbb{R}^{d_{\text{model}} \times d_{\text{sim}}},
            \]
            after which a scalar compatibility score is computed with a similarity function \footnote{Since the precision matrix must be symmetric, non-symmetric similarity functions require an explicit symmetrization step, such as $(J + J^T) / 2$.} for each edge \((i,j)\),
            \[
                J_{ij} = \text{Similarity}(s_i, s_j).
            \]
            
            The resulting interaction strengths are subsequently mapped into one of the walk-summable precision matrix constructions introduced in Section~3.2.1. The exact implementation of the learned and fixed precision matrices used in this paper in detail in Appendix section \ref{Appendix:Implementation_Details}.

        \subsubsection{Multi-Head Graph Linear Transformations}
        \label{sec:multihead}
            The independence and parameter sharing assumptions, introduced for scalability, may be an oversimplification of the system dynamics that we want to model. To add expressiveness back into the model, we propose multi-headed graph linear transformations for learned precision matrices, akin to multi-head attention in Transformers ~\cite{attn_all_u_need}. Through multi-headed Graph Linear Transformations, each head learns its own walk-summable topology $J^{(k)}$ and observation vector $h^{(k)}$, and performs a graph linear transformation independently to obtain head-specific graph linear transformation results. This formulation allows different heads to focus on distinct structural, semantic, or confidence contexts within the graph. The resulting node embeddings are layer-normalized, activated, concatenated, and projected back to the feature dimension. 
    
    \subsection{Addressing the Unbounded Memory}
        \label{sec:memory_efficiency}
        Although GaBP guarantees convergence under walk-summability, the number of iterations required to reach convergence is unknown. This poses a challenge for training, as unrolling the iterative inference process would require storing, potentially, an unbounded number of intermediate messages for each layer. To address this, we adopt an implicit differentiation strategy that computes exact gradients at convergence without unrolling the iterations. At convergence, the graph linear transformation satisfies \(J\mu = h\). Differentiating with respect to the model parameters
        \(\theta\) yields
        \[
            J\,\frac{d\mu}{d\theta} + \frac{dJ}{d\theta}\mu = \frac{dh}{d\theta}
            \;\Rightarrow\;
            J\,\frac{d\mathcal{L}}{dh} = \frac{d\mathcal{L}}{d\mu}.
        \]
        This relation allows gradients to be computed directly at the fixed point using the same GaBP solver as in the forward pass, eliminating the need to store intermediate messages. This reduces memory complexity from \(O(T E)\) to \(O(E + N)\), where \(T\) is the number of GaBP iterations, \(E\) the number of edges, and \(N\) the number of nodes, trading additional compute for substantial memory savings.

%%%%%%%%%%%%%%%%%%%%%%%%%%%%%%%%%%%%%%%%%%

\section{Experimental Setup}
    We evaluate the performance of learned and fixed topologies for graph linear transformation models across six benchmark datasets with cyclic graph structures under both homogeneous and heterogeneous learning settings. The Cora, Citeseer, and PubMed citation datasets serve as representative homogeneous benchmarks, while the Texas, Wisconsin, and Cornell datasets serve as representative heterogenous benchmarks. These six datasets each represent a single, static graph and are used for transductive node classification. 

    We compare our method against foundational and representative models from dynamic neighborhood aggregation methods, along with standard local aggregation methods. Specifically, we include the Graph Convolutional Network (GCN) \cite{GCN}, Graph Attention Network (GAT) \cite{GAT}, and GraphSage \cite{graphsage} as representative local aggregation models. For methods that perform global or continuous neighborhood aggregation, we evaluate against the Structure-Aware Transformer (SAT) \cite{structure_aware_transformer}, an adapted Graphormer \cite{graphormer} designed for node classification rather than graph classification, as well as the Graph Neural Differential Equation (GDE) \cite{GDE} and Continuous Graph Neural Network (CGNN) \cite{CGNN}. 

    To ensure a fair comparison across models, we fix the hidden feature dimension to 64 across all methods. For models based on stacked aggregation layers, we use two aggregation layers. For head-based architectures, each layer employs eight heads in the first layer and a single head in the second layer. Each first layer head outputs eight features, while the second layer head outputs all 64 features. Continuous-depth models do not admit a discrete notion of layer depth; for these models, we parameterize the vector field in the same 64-dimensional hidden space. The goal of this setup is to control for model capacity so the differences primarily reflect the behavior of each aggregation mechanism rather than variations in network depth or parameter count.

    All models are trained using the Adam optimizer with weight decay $5\cdot10^{-4}$, early stopping with a patience of 100 epochs and a dropout of $0.6$, except the fixed Laplacian which used a patience of 200 epochs. The learning rate for the graph linear transformation models and SAT models are $1\cdot 10^{-3}$ while the learning rate for the other models are $1\cdot10^{-2}$. The tolerance for the GaBP was set to $1\cdot10^{-6}$ and a maximum iteration count was set to 1000. To ensure stable convergence on graphs with loops, we apply message damping with a coefficient of $\lambda = 0.5$. For more information on message damping, we provide the pseudocode algorithm for GaBP in the Appendix section \ref{Appendix:pseudocode}. 
    
    For each dataset, we adopt random node splits preserving the original train/validation/test ratios but do not enforce class balance. This setup better reflects realistic scenarios where labeled nodes are unevenly distributed across classes, providing a more robust measure of generalization independent of split fragility \cite{shchur2019pitfalls}. Each experiment is repeated across 30 random runs, each with independently sampled splits. We report the mean $\pm$ standard deviation of test accuracy.

    For graph linear transformations, we ensure the graph is undirected and self loops are removed since the diagonal entries of the precision matrix $J$ already encode each node’s self-precision. For the other models, we ensure self loops. 

\section{Results}
    Table \ref{tab:results} reports test accuracy across the six benchmark datasets. Overall, the graph linear transformation models outperform or are on par to all baseline methods. However, learning topology is not always beneficial; the performance of the precision matrix depends on their respective inductive bias. 

   \begin{table*}[h]
        \centering
        \caption{The test accuracy in $\%$ evaluation datasets. \textbf{Bold} numbers indicate the best performance, while \underline{underline} indicate second best performance. “OOM” denotes out-of-memory errors encountered during training. (L) denotes a learned precision matrix while (F) denotes a fixed precision matrix.}
        \label{tab:results}
        \resizebox{0.9\textwidth}{!}{
            \begin{tabular}{lccccccc}
                \toprule
                  & Texas & Wisconsin & Cornell & Cora & Citeseer & Pubmed\\ 
                \textit{Homophily} & 0.11 & 0.21 & 0.30 & 0.81 & 0.74 & 0.80 \\

                \midrule
                GCN                 & 60.6$\pm$5.3          &          61.0$\pm$0.6 &          62.5$\pm$0.9 &          80.2$\pm$0.3 & \underline{65.3$\pm$0.6} & \underline{75.6$\pm$0.3} \\ 
                GAT                 & 48.6$\pm$0.0          &          64.5$\pm$2.5 &          63.4$\pm$5.5 &          76.6$\pm$1.1 &          64.6$\pm$1.8 &          74.0$\pm$0.5 \\ 
                GraphSage           & 78.4$\pm$0.0          & \textbf{85.0$\pm$1.0} &          76.9$\pm$1.8 &          76.9$\pm$0.4 &          63.4$\pm$1.6 &          71.0$\pm$0.4 \\    
                SAT                 & 77.8$\pm$2.6          &          82.7$\pm$2.4 &          77.9$\pm$2.6 &          53.5$\pm$1.4 &          47.6$\pm$2.2 &                   OOM \\
                Graphormer          & 69.4$\pm$7.6          &          65.2$\pm$4.3 &          57.7$\pm$7.0 &          31.7$\pm$2.3 &          28.7$\pm$3.4 &                   OOM \\
                GDE                 & 57.8$\pm$6.2          &          55.9$\pm$1.6 &          60.0$\pm$2.9 &          80.1$\pm$1.3 &          65.1$\pm$1.6 &  75.0$\pm$0.9\\
                CGNN                & 52.3$\pm$1.7          &          75.2$\pm$1.8 &          66.8$\pm$1.2 &          73.9$\pm$0.2 &          65.2$\pm$0.5 &          74.9$\pm$0.4 \\
                \midrule
                % ---- Learned block ----
                \multicolumn{7}{l}{\textbf{Ours}}\\
                (L) Pairwise Normal   & 82.1$\pm$4.2 &  78.1$\pm$2.2 &          75.5$\pm$4.3 &          46.4$\pm$1.2 &          39.0$\pm$1.2 &          64.1$\pm$1.3 \\
                (L) Diagonally Dominant & \textbf{83.7$\pm$3.7} &          76.5$\pm$1.8 &          74.9$\pm$4.6 &          47.7$\pm$1.3 &          41.9$\pm$1.5 &          62.3$\pm$1.5 \\ 
                (L) Laplacian       & 61.1$\pm$7.3          &          76.1$\pm$3.1 &          63.7$\pm$3.9 & \underline{80.4$\pm$1.3} &          64.5$\pm$1.3 &          74.2$\pm$1.4 \\
                (F) Pairwise Normal & \underline{82.7$\pm$3.4} & 76.8$\pm$1.9 & \underline{80.2$\pm$4.0} & 44.8$\pm$1.0 & 39.4$\pm$1.4 & 66.0$\pm$1.7\\
                (F) Diagonally Dominant & 56.2$\pm$7.0 & 68.4$\pm$3.0 & 57.0$\pm$5.2 & \textbf{80.6$\pm$1.2} & 65.1$\pm$1.4 & 75.3$\pm$0.9 \\
                (F) Laplacian        & 77.8$\pm$10.3 & \underline{83.1$\pm$2.0} & \textbf{81.3$\pm$2.3} & 71.9$\pm$1.4 & \textbf{66.0$\pm$1.8} & \textbf{76.8$\pm$1.0}\\
                \bottomrule
            \end{tabular}
        }
    \end{table*}
        Pairwise-normal constructions perform best on heterogeneous datasets, where selectively attending to informative edges mitigates the effects of heterophilic neighborhoods. In homophilic datasets, however, this selectivity restricts information flow, leading to reduced performance relative to less selective propagation mechanisms. 
        
        Laplacian precision matrices demonstrate the most stable performance across homophily levels, indicating that strong structural smoothing provides a robust prior for long-range information flow. Both fixed and learned Laplacian variants perform competitively on homophilic benchmarks while remaining effective in heterogeneous settings where local neighborhoods are unreliable, highlighting the benefits of community-level diffusion over aggressive edge filtering.
        
        Diagonally dominant constructions exhibit a sharper trade-off between homophilic and heterogeneous graphs, with performance strongly dependent on whether node confidence is learned or fixed. Learned confidence mechanisms enable selective regulation of information flow, improving performance on heterogeneous datasets by suppressing misleading neighborhoods, but this same selectivity interferes with the uniform smoothing required for homophilic graphs, leading to degraded accuracy. In contrast, fixed diagonally dominant matrices align naturally with homophilic structure, where uniform confidence supports consistent feature diffusion but fail to correct erroneous propagation paths in heterogeneous graphs. 
        
        These behaviors are reflected in the correlation structures induced by each precision matrix (Appendix \ref{Appendix:Analysis}): Laplacian constructions yield community-level correlation patterns, diagonally dominant matrices emphasize node centric influence, and pairwise-normal models induce highly selective pathways, collectively explaining the performance trends observed in Table 1.

    \subsection{Convergence Behavior}
        Table \ref{tab:iteration_results} reports the typical convergence iteration ranges of GaBP during training for each precision matrix. Because graph linear transformations rely on iterative inference, convergence behavior directly affects both computational efficiency and training dynamics.

       \begin{table}[h]
            \centering
            \caption{Typical GaBP iteration ranges required for forward and backward passes across precision matrix constructions. Values report observed ranges across datasets and training epochs.}
            \label{tab:iteration_results}
            \resizebox{\columnwidth}{!}{
                \begin{tabular}{lccc}
                    \toprule
                      & Forward Pass & Backward Pass\\ 
                    \midrule
                    (L) Pairwise Normal     & 20 - 40  & 15 - 25 \\
                    (L) Diagonally Dominant & 25 - 100 & 20 - 40 \\ 
                    (L) Laplacian           & 45 - 65  & 15 - 25 \\
                    (F) Pairwise Normal     & 20 - 30  & 15 - 20 \\
                    (F) Diagonally Dominant & 70 - 200 & 20 - 25 \\
                    (F) Laplacian           & $>$1000   & $>$1000    \\
                    \bottomrule
                \end{tabular}
            }
        \end{table}

        Two clear trends emerge. First, except for the fixed Laplacian, the backward pass consistently converges faster than the forward pass across all constructions. This suggests that gradient propagation occurs over a smoother signal than the raw node observations used during forward inference.
        
        Second, the fixed Laplacian is the only construction that does not converge within the allotted iteration budget, resulting in partially converged node mixing during both prediction and gradient computation. Despite this, the fixed Laplacian achieves the strongest overall performance. This indicates that full convergence may not be required for effective representation learning.

\section{Conclusion}
    This work introduced Graph Linear Transformations, an approach for incorporating direct and indirect contexts in graphs. By interpreting graphs as walk-summable Gaussian graphical models, we are able to perform a linear transformation on the graph according to its topology. Across both homophilic and heterophilic benchmarks, graph linear transformations consistently outperforms all global aggregation methods, while matching or surpassing local neighborhood methods. 

    Two key challenges to broader adoption are improving the efficiency of graph linear transformations and expanding the class of Gaussian graphical models that can be represented on arbitrary graph structures. Since graph linear transformations rely on iterative methods, there is a clear path toward scalability through well-established acceleration techniques, such as multigrid methods \cite{multigrid}, preconditioning, and hierarchical solvers. Moreover, walk-summable graphs represent only a subset of valid Gaussian graphical models. Existing extensions of GaBP, such as those in \cite{GaBP_WalkSum1}, demonstrate that inference can be performed on graphs that are not inherently walk-summable. We believe these directions offer promising avenues for future research and position graph linear transformations as a compelling alternative to both depth-based and dynamic aggregation methods.

    The code is available at \url{https://github.com/Marshall-Rosenhoover/Graph-Linear-Transformations}

\section*{Impact Statement}
    This paper presents work whose goal is to advance the field of Machine Learning. There are many potential societal consequences of our work, none which we feel must be specifically highlighted here.

\bibliography{references}

@inproceedings{
    GAT,
    title={Graph Attention Networks},
    author={Petar Veličković and Guillem Cucurull and Arantxa Casanova and Adriana Romero and Pietro Liò and Yoshua Bengio},
    booktitle={International Conference on Learning Representations},
    year={2018},
    url={https://openreview.net/forum?id=rJXMpikCZ},
}

@inproceedings{
    GCN,
    title={Semi-Supervised Classification with Graph Convolutional Networks},
    author={Thomas N. Kipf and Max Welling},
    booktitle={International Conference on Learning Representations},
    year={2017},
    url={https://openreview.net/forum?id=SJU4ayYgl}
}

@inproceedings{CGNN,
  title={Continuous Graph Neural Networks},
  author={Louis-Pascal Xhonneux and Meng Qu and Jian Tang},
  booktitle = {Proceedings of the 37th International Conference on Machine Learning},  
  year={2020},
  url={https://proceedings.mlr.press/v119/xhonneux20a.html}
}

@article{GDE,
  title={Graph Neural Ordinary Differential Equations},
  author={Poli, Michael and Massaroli, Stefano and Park, Junyoung and Yamashita, Atsushi and Asama, Hajime and Park, Jinkyoo},
  journal={arXiv preprint arXiv:1911.07532},
  year={2019}
}

@inproceedings{graphormer,
    author = {Ying, Chengxuan and Cai, Tianle and Luo, Shengjie and Zheng, Shuxin and Ke, Guolin and He, Di and Shen, Yanming and Liu, Tie-Yan},
    booktitle = {Advances in Neural Information Processing Systems},
    editor = {M. Ranzato and A. Beygelzimer and Y. Dauphin and P.S. Liang and J. Wortman Vaughan},
    pages = {28877--28888},
    publisher = {Curran Associates, Inc.},
    title = {Do Transformers Really Perform Badly for Graph Representation?},
    url = {https://proceedings.neurips.cc/paper_files/paper/2021/file/f1c1592588411002af340cbaedd6fc33-Paper.pdf},
    volume = {34},
    year = {2021}
}

@InProceedings{structure_aware_transformer,
  title = 	 {Structure-Aware Transformer for Graph Representation Learning},
  author =       {Chen, Dexiong and O'Bray, Leslie and Borgwardt, Karsten},
  booktitle = 	 {Proceedings of the 39th International Conference on Machine Learning},
  pages = 	 {3469--3489},
  year = 	 {2022},
  editor = 	 {Chaudhuri, Kamalika and Jegelka, Stefanie and Song, Le and Szepesvari, Csaba and Niu, Gang and Sabato, Sivan},
  volume = 	 {162},
  series = 	 {Proceedings of Machine Learning Research},
  month = 	 {17--23 Jul},
  publisher =    {PMLR},
  url = 	 {https://proceedings.mlr.press/v162/chen22r.html},
}

@inproceedings{GaBP_WalkSum1,
    author    = {Johnson, Jason K. and Bickson, Danny and Dolev, Danny},
    title     = {Fixing convergence of Gaussian belief propagation}, 
    booktitle = {2009 IEEE International Symposium on Information Theory},
    year      = {2009},
    volume    = {},
    number    = {},
    pages     = {1674-1678},
    doi       = {10.1109/ISIT.2009.5205777}}

@phdthesis{GaBP_Thesis,
    author = {Bickson, Danny},
    title  = {Gaussian Belief Propagation: Theory and Application},
    school = {The Hebrew University of Jerusalem},
    year   = {2008},
    url    = {https://www.cs.huji.ac.il/~dolev/pubs/thesis/phd-thesis-bickson.pdf}
}

@Article{malioutov2006walk,
    author  = {Dmitry M. Malioutov and Jason K. Johnson and Alan S. Willsky},
    title   = {Walk-Sums and Belief Propagation in Gaussian Graphical Models},
    journal = {Journal of Machine Learning Research},
    year    = {2006},
    volume  = {7},
    number  = {73},
    pages   = {2031--2064},
    url     = {http://jmlr.org/papers/v7/malioutov06a.html}
}

@inproceedings{prenorm,
    title     = {On Layer Normalization in the Transformer Architecture},
    author    = {Xiong, Ruibin and Yang, Yunchang and He, Di and Zheng, Kai and Zheng, Shuxin and Xing, Chen and Zhang, Huishuai and Lan, Yanyan and Wang, Liwei and Liu, Tieyan},
    booktitle = {Proceedings of the 37th International Conference on Machine Learning},
    pages     = {10524--10533},
    year      = {2020},
    editor    = {III, Hal Daumé and Singh, Aarti},
    volume    = {119},
    series    = {Proceedings of Machine Learning Research},
    month     = {13--18 Jul},
    publisher = {PMLR},
    url       = {https://proceedings.mlr.press/v119/xiong20b.html},
}

@inproceedings{sum_of_walks_intractable,
    author    = {Borgwardt, K.M. and Kriegel, H.P.},
    booktitle = {Fifth IEEE International Conference on Data Mining (ICDM'05)}, 
    title     = {Shortest-path kernels on graphs}, 
    year      = {2005},
    volume    = {},
    number    = {},
    pages     = {8 pp.-},
    doi={10.1109/ICDM.2005.132}
}

@InProceedings{sum_of_walks_intractable_2,
    author    = {G{\"a}rtner, Thomas and Flach, Peter and Wrobel, Stefan},
    editor    = {Sch{\"o}lkopf, Bernhard and Warmuth, Manfred K.},
    title     = {On Graph Kernels: Hardness Results and Efficient Alternatives},
    booktitle = {Learning Theory and Kernel Machines},
    year      = {2003},
    publisher = {Springer Berlin Heidelberg},
    address   = {Berlin, Heidelberg},
    pages     = {129--143},
    isbn      = {978-3-540-45167-9}
}

@article{GaBP_intro,
  title   = {A visual introduction to Gaussian Belief Propagation},
  author  = {Ortiz, Joseph and Evans, Talfan and Davison, Andrew J.},
  journal = {arXiv preprint arXiv:2107.02308},
  year    = {2021},
}

@inproceedings{attn_all_u_need,
    author    = {Vaswani, Ashish and Shazeer, Noam and Parmar, Niki and Uszkoreit, Jakob and Jones, Llion and Gomez, Aidan N and Kaiser, \L ukasz and Polosukhin, Illia},
    booktitle = {Advances in Neural Information Processing Systems},
    editor    = {I. Guyon and U. Von Luxburg and S. Bengio and H. Wallach and R. Fergus and S. Vishwanathan and R. Garnett},
    pages     = {},
    publisher = {Curran Associates, Inc.},
    title     = {Attention is All you Need},
    url       = {https://proceedings.neurips.cc/paper_files/paper/2017/file/3f5ee243547dee91fbd053c1c4a845aa-Paper.pdf},
    volume    = {30},
    year      = {2017}
}

@book{multigrid,
    author    = {Briggs, William and Henson, Van and McCormick, Steve},
    year      = {2000},
    month     = {01},
    pages     = {},
    title     = {A Multigrid Tutorial, 2nd Edition},
    isbn      = {978-0-89871-462-3},
    publisher = {SIAM: Society for Industrial and Applied Mathematics}
}

@misc{shchur2019pitfalls,
    title={Pitfalls of Graph Neural Network Evaluation}, 
    author={Oleksandr Shchur and Maximilian Mumme and Aleksandar Bojchevski and Stephan Günnemann},
    year={2019},
    eprint={1811.05868},
    archivePrefix={arXiv},
    primaryClass={cs.LG},
    url={https://arxiv.org/abs/1811.05868}, 
}

@inproceedings{graphsage,
    author    = {Hamilton, Will and Ying, Zhitao and Leskovec, Jure},
    booktitle = {Advances in Neural Information Processing Systems},
    editor    = {I. Guyon and U. Von Luxburg and S. Bengio and H. Wallach and R. Fergus and S. Vishwanathan and R. Garnett},
    pages     = {},
    publisher = {Curran Associates, Inc.},
    title     = {Inductive Representation Learning on Large Graphs},
    url       = {https://proceedings.neurips.cc/paper_files/paper/2017/file/5dd9db5e033da9c6fb5ba83c7a7ebea9-Paper.pdf},
    volume    = {30},
    year      = {2017}
}
\bibliographystyle{icml2026}

%%%%%%%%%%%%%%%%%%%%%%%%%%%%%%%%%%%%%%%%%%%%%%%%%%%%%%%%%%%%%%%%%%%%%%%%%%%%%%%
%%%%%%%%%%%%%%%%%%%%%%%%%%%%%%%%%%%%%%%%%%%%%%%%%%%%%%%%%%%%%%%%%%%%%%%%%%%%%%%
% APPENDIX
%%%%%%%%%%%%%%%%%%%%%%%%%%%%%%%%%%%%%%%%%%%%%%%%%%%%%%%%%%%%%%%%%%%%%%%%%%%%%%%
%%%%%%%%%%%%%%%%%%%%%%%%%%%%%%%%%%%%%%%%%%%%%%%%%%%%%%%%%%%%%%%%%%%%%%%%%%%%%%%
\newpage
\appendix
\onecolumn

\section{Appendix}
    \subsection{Implementation details of the Precision Matrices}
    \label{Appendix:Implementation_Details}
        The following paragraphs describe our implementations of each precision matrix. These paragraphs directly describe their corresponding precision matrix implementation in our github. However, we want to note that these are not the only way to implement the precision matrices, and we encourage the research into more well designed precision matrices.
        
        \paragraph{Pairwise Normal Fixed}
            In standard Gaussian graphical models, the observation vector was typically used for defining the edges. Likewise for the fixed pairwise normal, we use the base input features to define the fixed pairwise normal matrix. The cosine similarity was the similarity function used to determine the strength of edges, $b_{ij}$, which was scaled to be between $[-0.99, 0.99]$. The confidence for each node for each edge was set to 1 such that the pairwise constraint is always satisfied. Lastly, the matrix is symmetrically normalized.
            
        \paragraph{Pairwise Normal Learned}
            In standard Gaussian graphical models, the observation vector was typically used for defining the edges. In this case, we use the learned observation vector to define our edge interactions. For the edge confidences, we let each node learn an unbounded self confidence for each edge pair. Thus, for an edge $i \to j$, we use a linear layer on the concatenation of the learned observation vector from both nodes: $a_{ij} = softplus([h_i||h_j]) + 1\cdot10^{-8}$ and $c_{ij} = softplus([h_j||h_i]) + 1\cdot10^{-8}$. To define the limits of the edge strength, we produce a scalar to scale the similarity function $scale = \epsilon\sqrt{a_{ij} c_{ij}}$, where $\epsilon$ is between [0, 1). We then scale the similarity function, cosine similarity in this case, using the scalar. Lastly, the matrix is symmetrically normalized. 

        \paragraph{Diagonally Dominant Fixed}
            For the fixed Diagonally Dominant matrix, we assume that each edge has a strength of 1. For the confidence of each node, we assume it is 1 plus the degree of the node. 
        
        \paragraph{Diagonally Dominant Learned}
            For the learned Diagonally Dominant matrix, we reuse the learned observation vector for the similarity vector. For the similarity function, we use the cosine similarity. For the learned self confidence, we take the original node embeddings $X$ and use them to produce node unbounded specific confidences: $J_{ii} = \text{Softplus}(X W)$. We then symmetric normalize the matrix.
        \paragraph{Laplacian Fixed}
            For the fixed Laplacian, we take the normal adjacency matrix and have the diagonal be $1\cdot10^{-8}$ + the degree of the node. We then symmetrically normalize the matrix and scale the off diagonals by 0.99 so the eigenvalues are between [-0.99, 0.99].
            
        \paragraph{Laplacian Learned}
            For the learned Laplacian, we use a separate observation vector and similarity vector. Edge weights are defined using a Gaussian kernel on the learned similarity embeddings, producing nonnegative edge strengths. These weights are degree-normalized to form a Laplacian-style precision matrix, where off-diagonal entries encode normalized negative interactions. To ensure walk-summability, the off-diagonal terms are scaled to lie strictly within $(-1, 1)$, and the diagonal entries consist of a fixed spectral floor plus a small learned node-specific confidence. The learned diagonal contribution is bounded to prevent it from dominating the Laplacian structure. Finally, the matrix is symmetrically normalized and then scaled by a learned global sigmoid factor, following the normalization principles of \cite{GCN}.

    \subsection{Pseudocode for Gaussian Belief Propagation}
    \label{Appendix:pseudocode}
        During GaBP, message damping can be beneficial for improving convergence stability. Damping replaces each message update with a convex combination of its previous value and the newly computed value, effectively acting as a moving average over iterations. This is particularly helpful on frustrated graphs, where competing edge interactions cannot be simultaneously satisfied and undamped message updates may oscillate.

        \begin{algorithm}[H]
            \caption{Gaussian Belief Propagation}
            \label{alg:GaBP}
            \begin{algorithmic}[1]
            \STATE \textbf{Input:} $J, h, \varepsilon, \lambda$ 
            \STATE Initialize $\pi_{i\to j}=0$, $\eta_{i\to j}=0$ for all $(i,j)\in E$
            \WHILE{not converged ($\Delta > \varepsilon$)}
                \STATE $\alpha_{i\setminus j} = J_{ii} + \sum_{k\in N(i)\setminus j}\pi_{k\to i}$
                \STATE $\beta_{i\setminus j} = h_i + \sum_{k\in N(i)\setminus j}\eta_{k\to i}$
                \STATE $\pi^{\text{new}}_{i\to j} = -J_{ij}^2 / \alpha_{i\setminus j}$
                \STATE $\eta^{\text{new}}_{i\to j} = -J_{ij}\,\beta_{i\setminus j} / \alpha_{i\setminus j}$
                \STATE $\Delta\pi_{i\to j} = \pi^{\text{new}}_{i\to j} - \pi_{i\to j}$
                \STATE $\Delta\eta_{i\to j} = \eta^{\text{new}}_{i\to j} - \eta_{i\to j}$
                \STATE $\pi_{i\to j} \leftarrow (1-\lambda)\,\pi_{i\to j} + \lambda\,\pi^{\text{new}}_{i\to j}$
                \STATE $\eta_{i\to j} \leftarrow (1-\lambda)\,\eta_{i\to j} + \lambda\,\eta^{\text{new}}_{i\to j}$
                \STATE $\Delta = \max_{(i,j)\in E} \left( \max\left( |\Delta\pi_{i\to j}|,\; |\Delta\eta_{i\to j}| \right) \right)$
            \ENDWHILE
            \STATE \textbf{Output:} $\mu_i = \eta_i / \pi_i$
            \end{algorithmic}
        \end{algorithm}

        Algorithm~\ref{alg:GaBP} provides the pseudocode for GaBP with message damping and serves as a reference for the equations presented in the background section.

    \subsection{Analysis of the Biases of Matrix Constructions}        
        \label{Appendix:Analysis}
\begin{figure}[t]
          \centering
        
          \begin{subfigure}{0.32\linewidth}
            \centering
            \includegraphics[width=\linewidth]{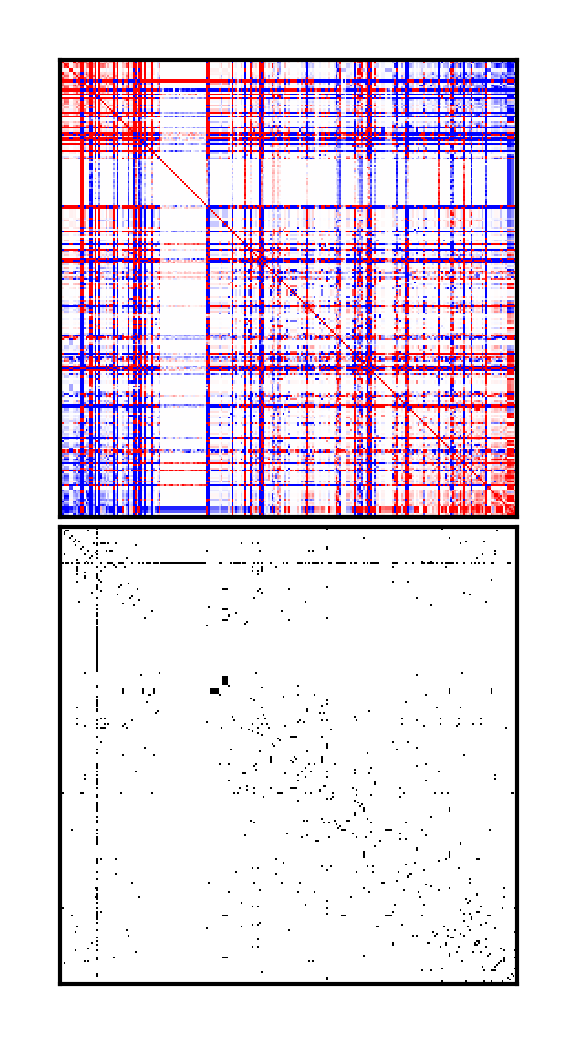}
            \caption{Fixed Laplacian}
            \label{fig:laplacian}
          \end{subfigure}
          \hfill
          \begin{subfigure}{0.32\linewidth}
            \centering
            \includegraphics[width=\linewidth]{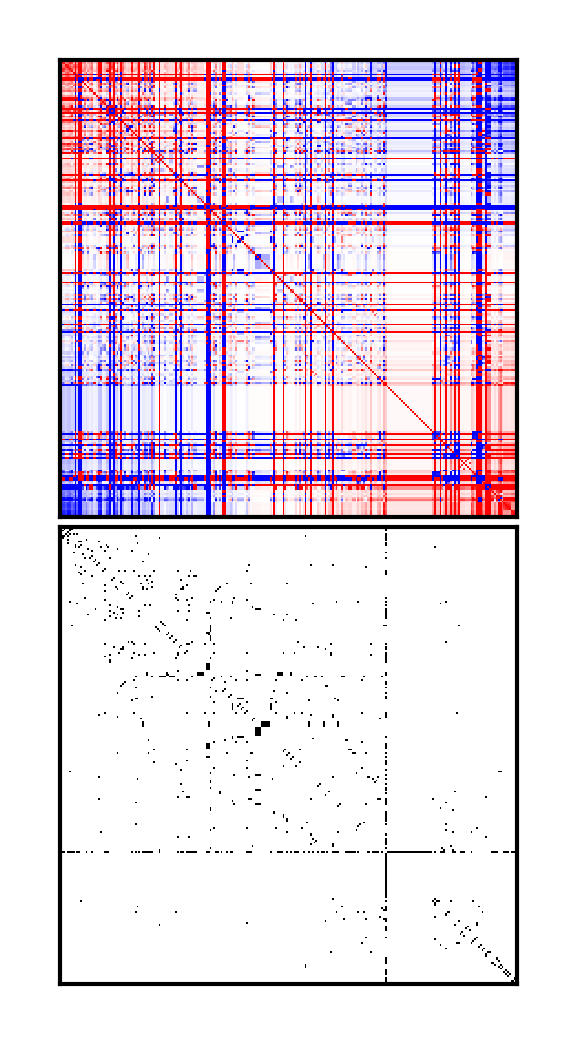}
            \caption{Fixed Diagonally Dominant}
            \label{fig:diagdom}
          \end{subfigure}
          \hfill
          \begin{subfigure}{0.32\linewidth}
            \centering
            \includegraphics[width=\linewidth]{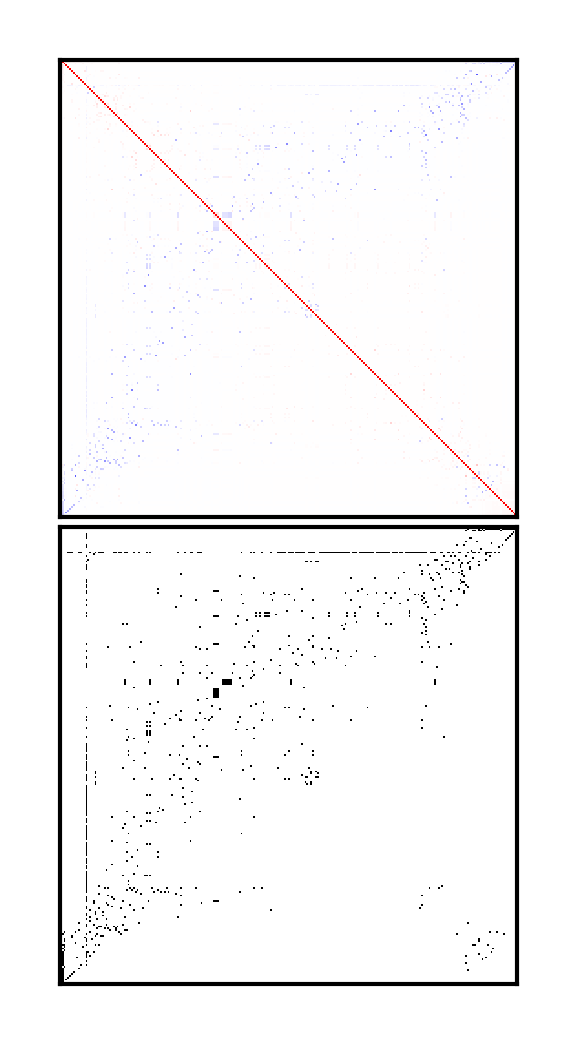}
            \caption{Fixed Pairwise Normal}
            \label{fig:pairwise}
          \end{subfigure}
        
          \caption{Comparison of fixed precision matrix constructions on the Wisconsin dataset.}
          \label{fig:precision-matrix-comparison}
        \end{figure}

\begin{figure}[h]
            \centering
            \includegraphics[width=\linewidth]{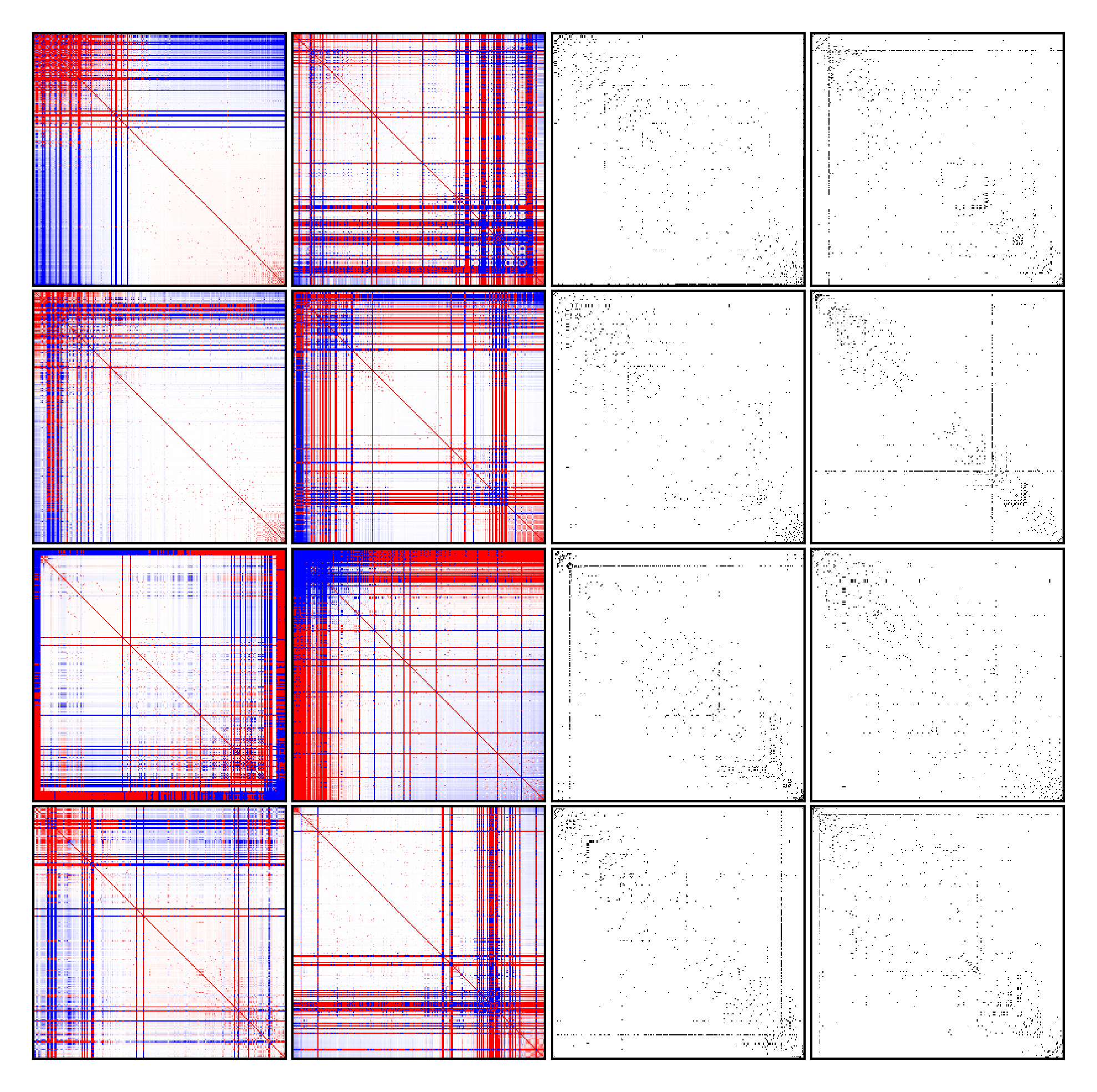}
            \caption{Learned Laplacian precision construction.}
            \label{fig:laplacian_wisconsin}
        \end{figure}

\begin{figure}[h]
            \centering
            \includegraphics[width=\linewidth]{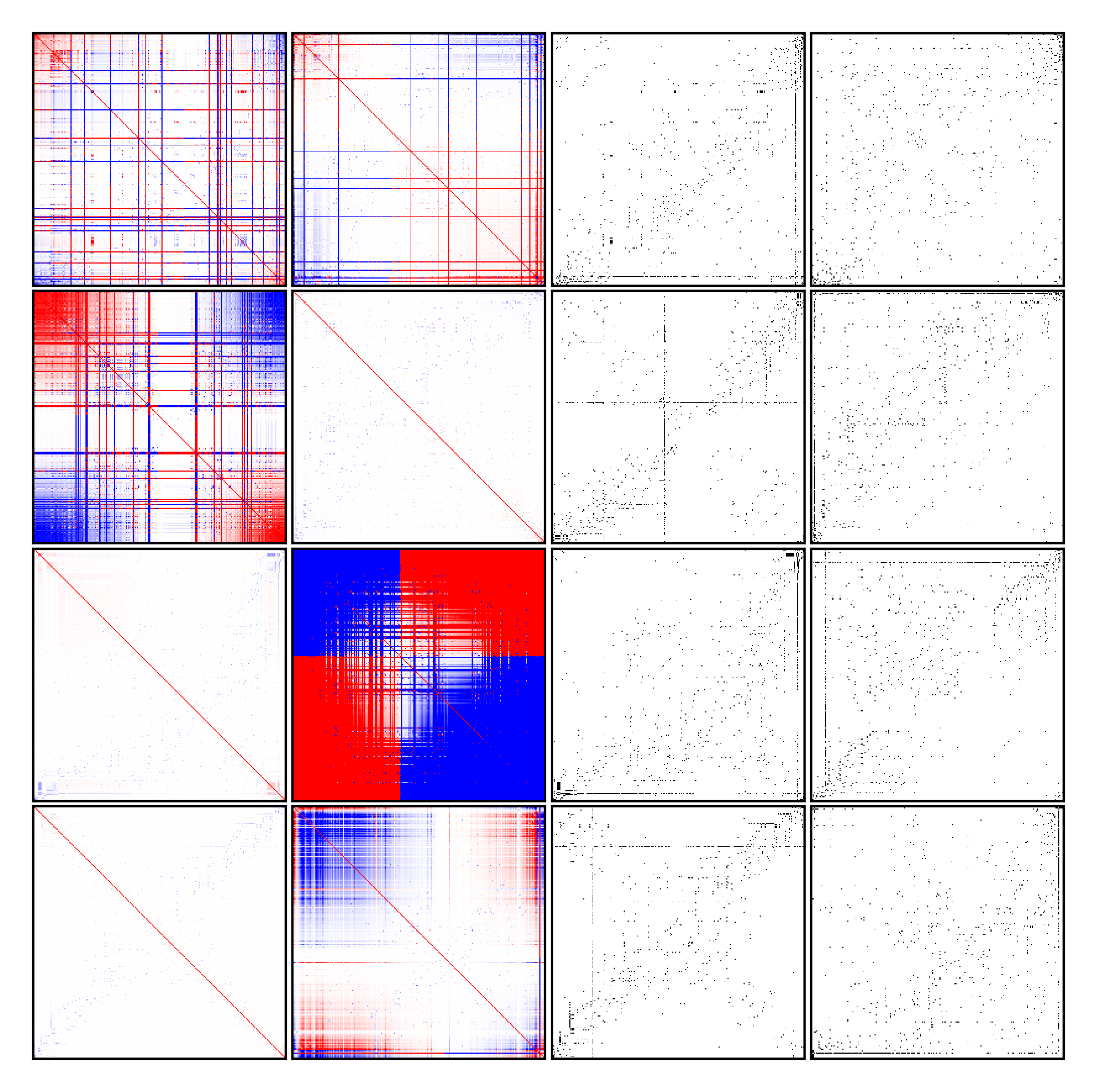}
            \caption{Learned Diagonally dominant precision construction.}
            \label{fig:dominant_wisconsin}
        \end{figure}

\begin{figure}[h]
            \centering
            \includegraphics[width=\linewidth]{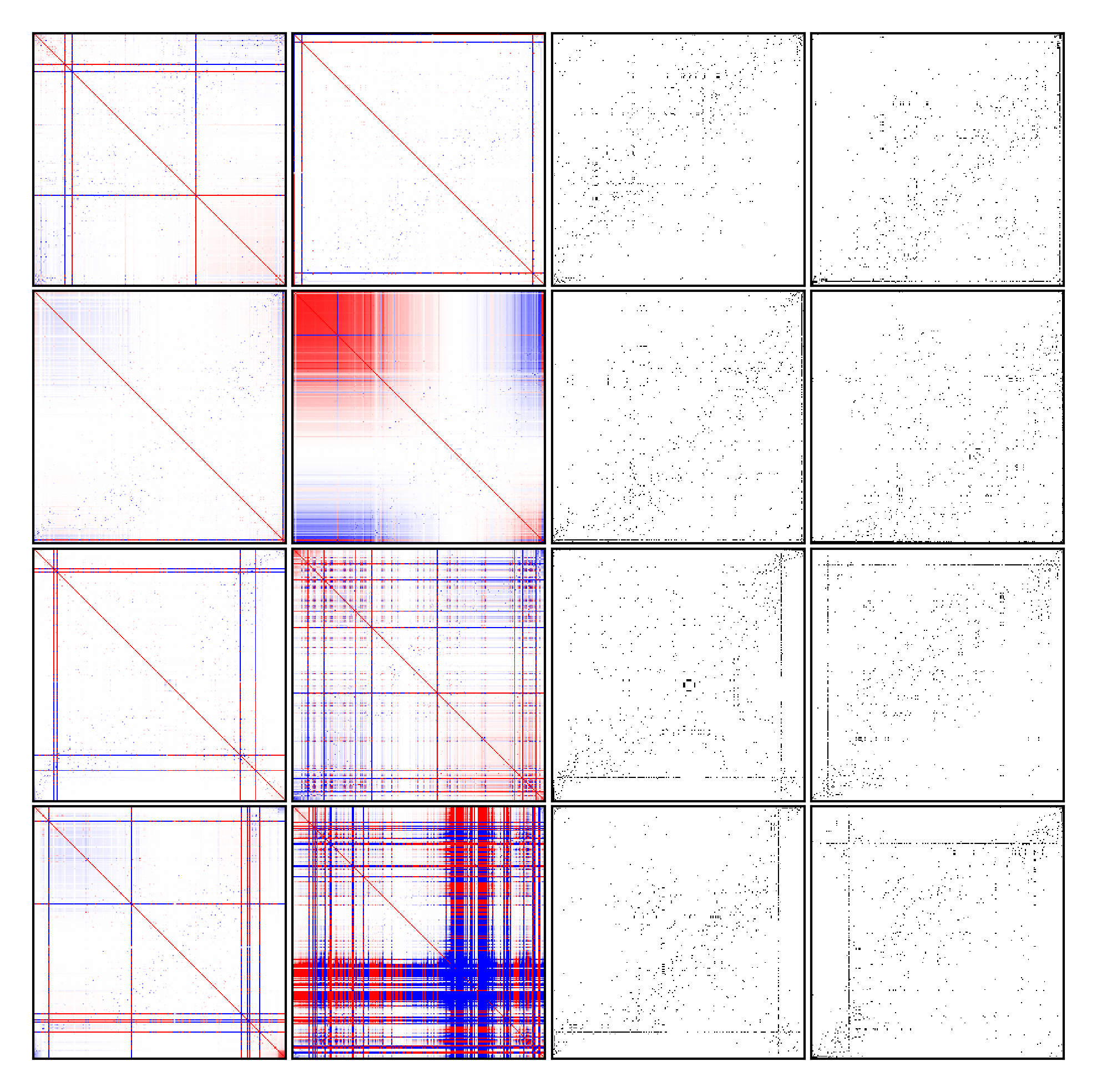}
            \caption{Learned Pairwise normal precision construction.}
            \label{fig:pairwise_wisconsin}
        \end{figure}

        To understand how each precision matrix construction shapes the topology, we visualize the correlation matrix (Figures~\ref{fig:precision-matrix-comparison}–\ref{fig:pairwise_wisconsin}) on the Wisconsin dataset. For the learned precision matrices, we visualize the first-layer. These correlation matrices are computed from the inverse precision matrix and describe how strongly two nodes would co-vary under the learned topology in the absence of any observation vector \(h\). They therefore characterize the influence structure implied by \(J\), meaning the potential pathways and relative strengths along which GaBP is capable of transmitting information, regardless of the actual evidence supplied during inference. Red values indicate positive correlations and blue values indicate negative correlations, with intensity proportional to magnitude. Nodes are ordered by the second eigenvector to highlight community structure, and the accompanying adjacency matrices (bottom panels for Figure ~\ref{fig:precision-matrix-comparison} and right panels for Figures \ref{fig:laplacian_wisconsin}-\ref{fig:pairwise_wisconsin}), ordered in the same way, show the corresponding learned connectivity patterns. 
        
        \paragraph{Fixed Constructions.}
            The fixed precision matrices show that the Laplacian and diagonally dominant are more homogenous in their representation, while the fixed pairwise normal is quite selective in its neighboring representations.

        \paragraph{Learned Laplacian Construction.}
            The Laplacian-based precision matrix produces a pronounced block structure resembling a plaid pattern of alternating positive and negative correlations. Nodes form compact clusters with internal correlations, while inter-cluster relationships often alternate in sign, indicating that the model captures both homophilic and heterophilic dependencies. The dense regions along the diagonal of the reordered adjacency matrix confirm the emergence of localized communities. Because the Laplacian construction constrains the eigenvalues of the precision matrix to lie within $(-1, 1)$, each node contributes with comparable influence to the overall propagation. Most of the learning therefore occurs through topological deformation, where the model folds the underlying space before blending node representations. As a result, this design behaves like a band-pass filter, supporting community-level diffusion over a learned relational manifold while maintaining structured contrast between clusters. This can be seen with the different block structures present in the adjacency matrices.

        \paragraph{Learned Diagonally Dominant Construction.}
            In the diagonally dominant formulation, a few high-confidence nodes can exert strong influence over their neighborhoods. Because each node’s self-term dominates its row in \( J \), information primarily flows outward from confident nodes rather than through collective averaging. Each node is equipped with a learned confidence gate that dynamically adjusts its self-precision, regulating how much information it absorbs from its neighbors. As a result, highly confident nodes act as local leaders. When node confidences are comparable, correlations weaken, leading to sparse yet focused connectivity patterns or minimal inter-node correlation. The more uniform edge spread, observed in the adjacency matrix, reflects this shift from community-level organization toward node-level importance. Overall, this design behaves as a low-pass filter centered around the most confident nodes.

        \paragraph{Learned Pairwise Normal Construction.}
            The pairwise normal matrix yields the most selective propagation pattern. Since each edge’s contribution depends on the mutual confidence of its connected nodes, information flows only where representations are jointly compatible. This formulation models each edge as the agreement both endpoints have in their connection, scaling similarity by their confidence ratio and emphasizing edges supported by reciprocal certainty. Consequently, propagation strength is highest along confident paths and suppressed in uncertain regions, requiring mutual trust for effective global communication. The resulting correlations form gradual gradients rather than sharp blocks, functioning as a high-pass filter in low-correlation areas and a low-pass filter in highly correlated areas. The corresponding adjacency matrix reveal clearer separation between sparse and dense zones, indicating that low-degree nodes propagate information more readily than high-degree ones, where competition among edges dampens transmission. 

%%%%%%%%%%%%%%%%%%%%%%%%%%%%%%%%%%%%%%%%%%%%%%%%%%%%%%%%%%%%%%%%%%%%%%%%%%%%%%%
%%%%%%%%%%%%%%%%%%%%%%%%%%%%%%%%%%%%%%%%%%%%%%%%%%%%%%%%%%%%%%%%%%%%%%%%%%%%%%%

\end{document}